\documentclass[letterpaper, 10 pt, conference]{ieeeconf}  

\usepackage{graphicx}
\usepackage{caption}
\usepackage{subcaption}
\usepackage{amsmath,amsfonts,amssymb}
\usepackage{array}
\usepackage{colortbl}
\usepackage{boldline}
\usepackage{algorithmic}
\usepackage[dvipsnames]{xcolor}
\usepackage{algorithm}
\usepackage{array}
\usepackage{textcomp}
\usepackage{stfloats}
\usepackage{url}
\usepackage{verbatim}

\usepackage[utf8]{inputenc} 
\usepackage[T1]{fontenc}
\usepackage{url}            

\usepackage{booktabs}
\usepackage{caption}

\usepackage{nicefrac}       
\usepackage{microtype}      
\usepackage{xcolor}      
\usepackage{graphicx}

\usepackage{mathtools}
\usepackage{tabularx}
\usepackage{framed}
\usepackage{bbm}
\usepackage{bm}
\usepackage{xspace}
\usepackage{amssymb}
\usepackage{multirow}
\usepackage{physics}
\usepackage{siunitx}
\usepackage{algorithm}
\usepackage{algorithmic}

\usepackage{cite}
\usepackage{tikz}
\usepackage{pifont}
\usepackage{easybmat}
\usepackage{multirow,bigdelim}
\usepackage{nccmath}

\usepackage{colortbl}

\usepackage{placeins}

\usepackage{subcaption}

\usepackage{lipsum}

\usepackage{thmtools}
\usepackage{thm-restate}
\usepackage{xcolor}

\makeatletter
\let\NAT@parse\undefined
\makeatother
\usepackage{hyperref}       \hypersetup{
    colorlinks,
    linkcolor={blue!90!black},
    citecolor={blue!90!black},
urlcolor={blue!95!black}
}

\usepackage[capitalise]{cleveref}       \usepackage{crossreftools}  \usepackage[all=normal,paragraphs=tight,floats=tight,mathspacing=tight,wordspacing=tight,tracking=tight]{savetrees} 
\usetikzlibrary{matrix,decorations.pathreplacing}
\usetikzlibrary{decorations.pathreplacing, positioning, calc}

\definecolor{Gray}{gray}{0.85}
\definecolor{LightCyan}{rgb}{0.88,1,1}
\definecolor{darkpurple}{RGB}{128,0,128}
\definecolor{darkbrown}{RGB}{51, 25, 0}
\definecolor{LightGray}{gray}{0.93}
\definecolor{mygreen}{rgb}{0.0, 0.5, 0.0}

\IEEEoverridecommandlockouts                              

\title{\LARGE \bf
ComTraQ-MPC: Meta-Trained DQN-MPC Integration for Trajectory Tracking with Limited Active Localization Updates
}

\author{Gokul Puthumanaillam$^{*}$, Manav Vora$^{*}$ and Melkior Ornik
\thanks{*Equal contribution}
\thanks{This work was supported by ONR grant N00014-23-1-2505.}
\thanks{Authors are with the Department of Aerospace Engineering and the Coordinated Science Laboratory, University of Illinois Urbana-Champaign,
Urbana, USA.
{\tt\small \{gokulp2,mkvora2,mornik\}}@illinois.edu}%
}

\begin{document}

\maketitle
\thispagestyle{empty}
\pagestyle{empty}


\begin{abstract}
Optimal decision-making for trajectory tracking in partially observable, stochastic environments where the number of active localization updates---the process by which the agent obtains its true state information from the sensors---are limited, presents a significant challenge. 
Traditional methods often struggle to balance resource conservation, accurate state estimation and precise tracking, resulting in suboptimal performance. 
This problem is particularly pronounced in environments with large action spaces, where the need for frequent, accurate state data is paramount, yet the capacity for active localization updates is restricted by external limitations.
This paper introduces ComTraQ-MPC, a novel framework that combines Deep Q-Networks (DQN) and Model Predictive Control (MPC) to optimize trajectory tracking with constrained active localization updates. The meta-trained DQN ensures adaptive active localization scheduling, while the MPC leverages available state information to improve tracking. The central contribution of this work is their reciprocal interaction: DQN's update decisions inform MPC's control strategy, and MPC's outcomes refine DQN's learning, creating a cohesive, adaptive system.
Empirical evaluations in simulated and real-world settings demonstrate that ComTraQ-MPC significantly enhances operational efficiency and accuracy, providing a generalizable and approximately optimal solution for trajectory tracking in complex partially observable environments.
\href{https://github.com/gokulp01/comtraq-mpc}{\textcolor{purple}{[Code]}}\footnote{\url{https://github.com/gokulp01/comtraq-mpc}}
\href{https://youtu.be/evpeBYR2GPE}{\textcolor{purple}{[Video]}}\footnote{\url{https://youtu.be/evpeBYR2GPE}}
\end{abstract}

\section{Introduction}
In the context of autonomous navigation, an active localization update \cite{lauri2022partially, joo2020autonomous} refers to an action which provides the agent with its true state information, from sensor data, within the environment it is operating in.
Addressing trajectory tracking within environments characterized by a limited number of active localization updates and partial observability presents a critical challenge in the development of autonomous systems \cite{leon2021review}.
Such constraints are not merely technical limitations but are often necessitated by the operational environment itself \cite{bailey2022design}. 
Every instance of actively localizing is a double-edged sword --- it helps the agent get better state estimates using sensory data at the cost of depletion of limited resources \cite{10136380} or increased risk of detection in sensitive applications \cite{ravela1994stealth}.
The traditional paradigms of trajectory tracking, heavily reliant on uninterrupted data flows for state estimation and control \cite{Nascimento_Dórea_Gonçalves_2018, kunhe2005mobile, 982266}, are ill-equipped to handle such scenarios.

The advent of learning-based methods \cite{kamran2019learning, xu2023drlbased, xu2020learning} has introduced significant advancements in autonomous trajectory tracking. However, these approaches often grapple with the complexities of large state and action spaces \cite{bharilya2024machine}, requiring substantial training and continuous dependency on true state data. Hence limited number of active localization updates and partial observability become a bottleneck in scenarios where the availability of such sensory data is heavily restricted \cite{schoellig2012optimization}.

Recent studies have explored deep reinforcement learning for optimal communication strategies in multi-agent systems \cite{zhu2024survey, mao2017accnet}. 
A common limitation among these works is that, they presuppose full observability of the agent's own state and partial observability of the other agents. Hence, the communication action is only used to estimate the states of other agents and not for localizing the agent itself. 

\begin{figure}[t]
    \centering
    \begin{subfigure}{0.23\textwidth}
        \centering
        \includegraphics[width=\linewidth, clip, trim=0 0 0 0]{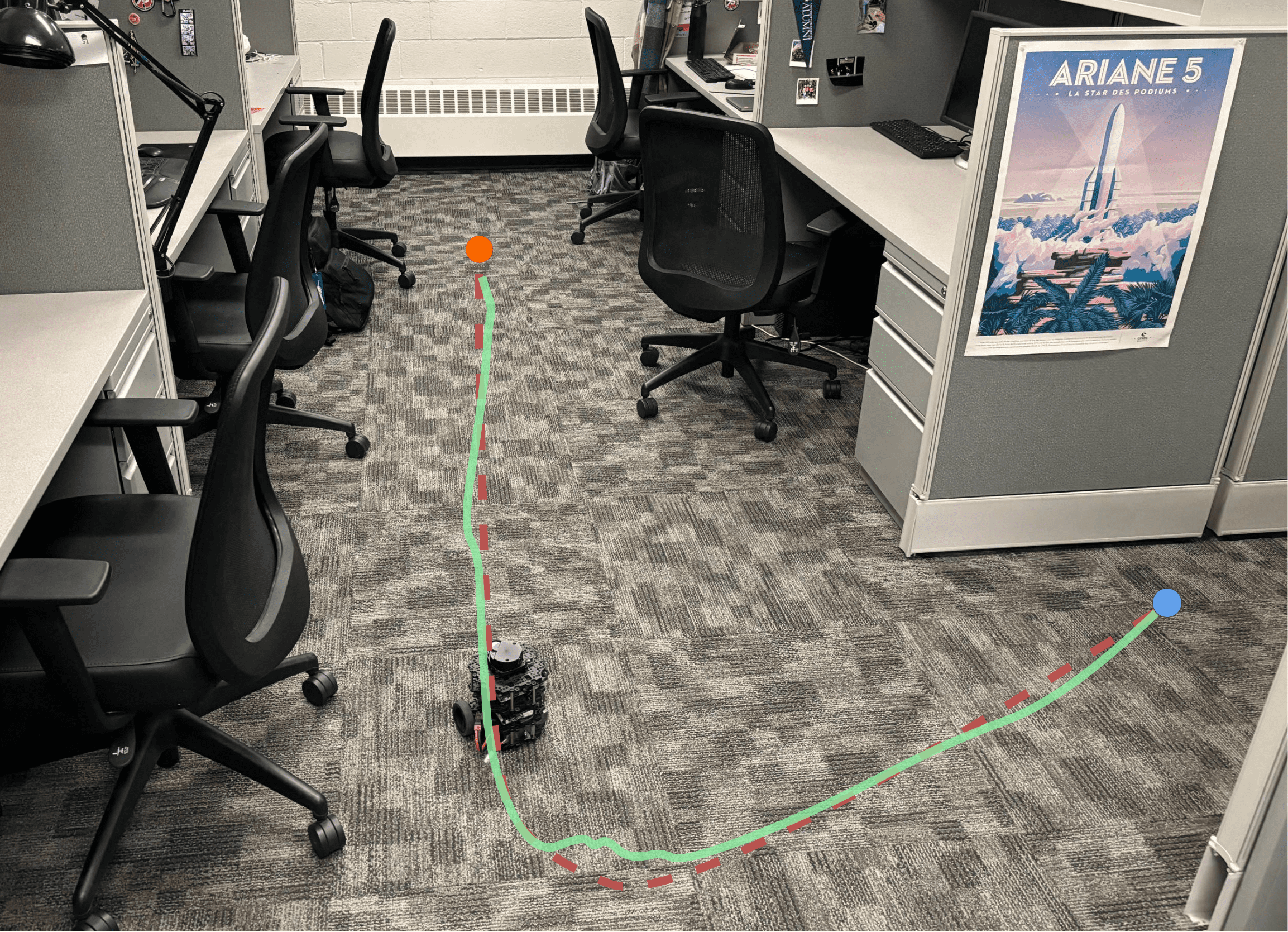} 
        \caption{ }
        \label{fig:teaser1}
    \end{subfigure}%
    \hfill
    \begin{subfigure}{0.24\textwidth}
        \centering
        \includegraphics[width=\linewidth, clip, trim=0 0 0 0]{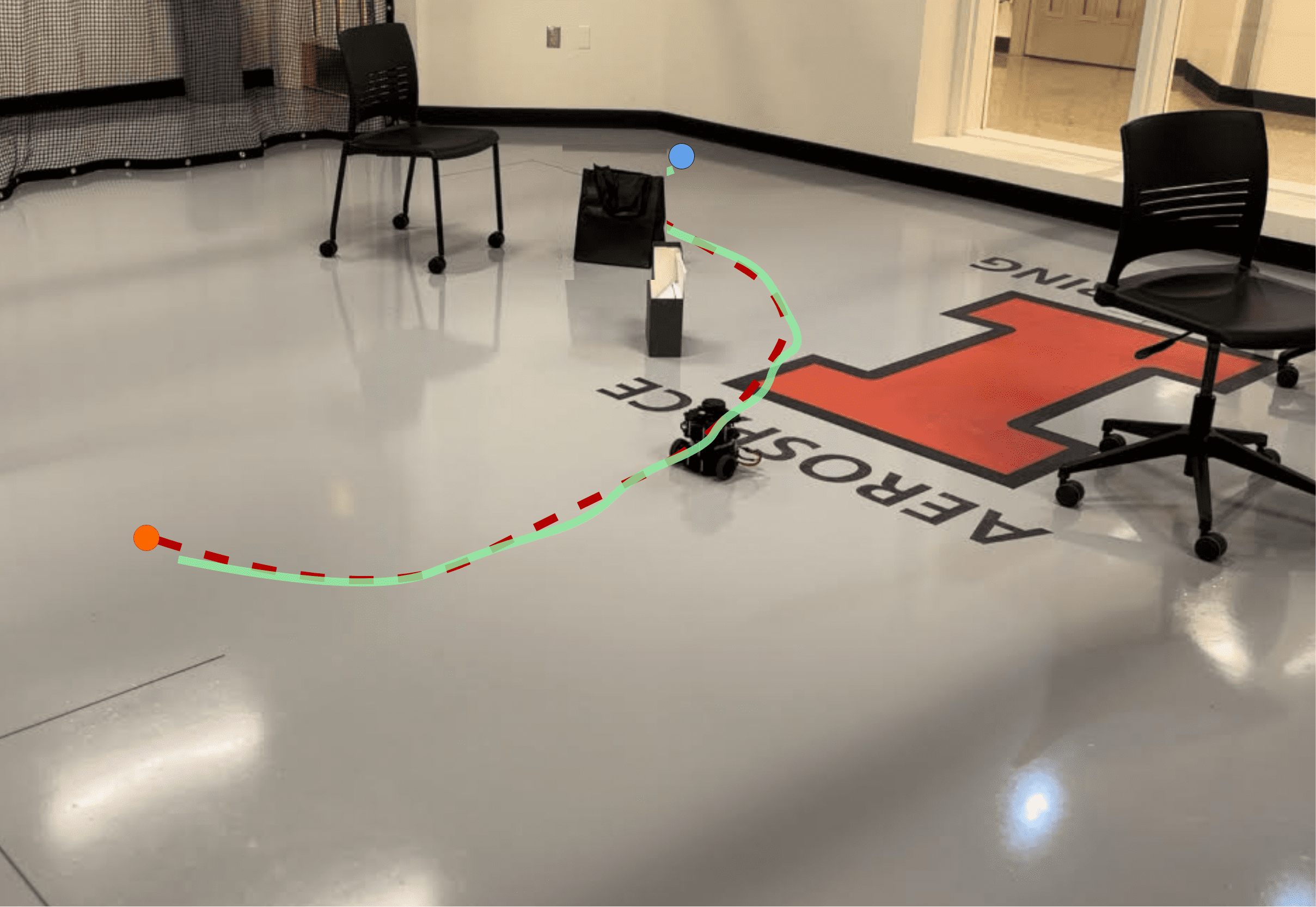} 
        \caption{}
        \label{fig:teaser2}
    \end{subfigure}%
    \caption{Performance of ComTraQ-MPC: Fig. (a) depicts ComTraQ-MPC in a known (previously seen during meta-training), trajectory (134 waypoints, active localization budget: 10) and (b) in a previously unseen, longer trajectory (245 waypoints, active localization budget: 20). \textcolor{Cyan}{Start} and \textcolor{orange}{goal} points are marked by \textcolor{Cyan}{blue} and \textcolor{orange}{orange} dots, respectively, with the \textcolor{BrickRed}{red} dotted line showing the \textcolor{BrickRed}{reference trajectory} and the \textcolor{LimeGreen}{green} line the \textcolor{LimeGreen}{ComTraQ-MPC path}.}
\end{figure}
Additionally, some studies have tackled the issue of the agent's own state's partial observability by framing the trajectory tracking problem as a Partially Observable Markov Decision Process (POMDP). The research in \cite{sun2018stochastic} introduces a tree search strategy for addressing trajectory tracking formulated as a POMDP, which may lead to computational challenges with expanding action spaces \cite{janson2017monte}. Likewise, \cite{van2017motion} employs belief space planning for navigating partially observable scenarios. 
However, these approaches are solely dependent on passive localization updates, utilizing merely the agent's belief state for planning purposes. Consequently, they neglect the significant benefits that could be derived from strategically leveraging sensor data to achieve precise state estimation via active localization updates.



The contribution of ComTraQ-MPC to this ongoing dialogue is significant, offering a comprehensive framework that not only learns when to perform an active localization update to access true state data, but also how to effectively integrate these decisions with trajectory planning and control strategies, as highlighted in our empirical evaluations.

\subsection{Contributions}
Recognizing the limitations of existing approaches, we formulate our problem as a Budgeted POMDP (b-POMDP) to facilitate strict adherence to budget constraints in environments with partial observability and limited active localization updates.  
The core contribution of our work lies in the novel integration of Deep Q-Networks (DQN) \cite{mnih2015human} with Model Predictive Control (MPC) \cite{garcia1989model}, forming the ComTraQ-MPC framework. 
The DQN component, meta-trained across diverse trajectories and budgets, brings to the framework an adaptive capability for making update decisions---evaluating the value of active localization based on the current state and trajectory of the agent. 
Conversely, the MPC component focuses on trajectory tracking, utilizing the information available at each decision point to track the reference trajectory.
The essence of our contribution is encapsulated in the bidirectional feedback mechanism between DQN and MPC.
 Localization decisions made by the DQN directly influence the quality of the state information that MPC uses for planning. Better-informed state estimates lead to more accurate trajectory tracking. In return, the outcomes of MPC's optimization provide a learning signal for the DQN in the form of the optimized MPC cost function. Through this signal, the DQN learns not only from the immediate outcomes of its decisions but also from the longer-term impacts on trajectory tracking efficiency as mediated by MPC, effectively closing the loop. 
 We test our approach in real-world settings and our results show improved trajectory tracking performance in environments with active localization update constraints compared to previously established baselines.


\section{Preliminaries}
We formulate the problem of trajectory tracking in environments with limited active localization updates as a special case of POMDPs with budget constraints. Hence we use the budgeted POMDP framework. 
\begin{figure*}[th]
    \centering
    \includegraphics[width=0.8\textwidth]{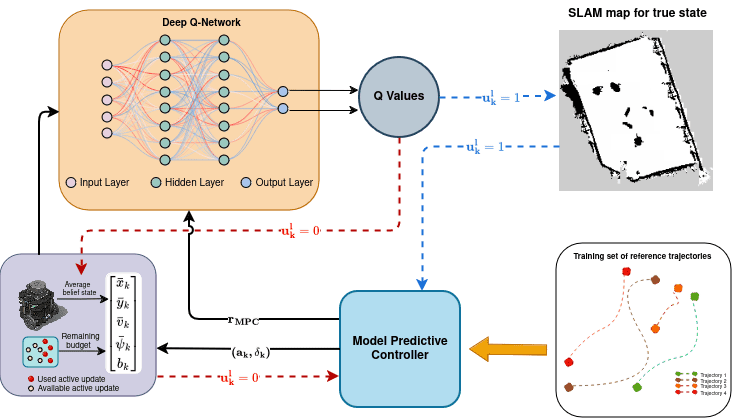}
    \caption{Architectural Overview of ComTraQ-MPC Framework. This diagram illustrates the integration of DQN for dynamic localization decision-making with MPC for precise trajectory tracking. The architecture encapsulates the synergy between adaptive active localization scheduling and robust trajectory tracking, highlighting the flow of information and decision processes that enable effective decision-making in environments with active localization update constraints.}
    \label{fig:arch}
\end{figure*}

\subsection{Budgeted POMDP}
A Budgeted Partially Observable Markov Decision Process (b-POMDP) \cite{10136380} in discrete time with a finite horizon and budget is formalized by an 9-tuple \( (S, A, T, \Omega, O, R, H, B, C(.) ) \). Here, \(S\) is a finite set of states; \(A\) is a finite set of actions; \(T: S \times A \rightarrow \Delta(S)\) is the transition probability function, where \(\Delta(S)\) indicates the set of probability distributions over \(S\); \(\Omega\) is a finite set of observations; \(O: \Omega \times S \times A \rightarrow \Delta(\Omega)\) is the observation probability function, defining the probability distribution over observations; \(R: S \times A \rightarrow [R_{\text{min}}, R_{\text{max}}]\) is the reward function, assigning a reward value to each state-action pair; \(H\) and \(B\) are elements of the set of non-negative integers, $\mathbb{N}_0$, specifying the finite planning horizon and the total budget for planning, respectively; $C(.)$ is the cost function which gives the cost incurred by performing an action $a \in A$.

At a discrete time step \(k\), the agent occupies state \(s_k \in S\). 
This state $s_k$ has two components: a fully observable cost component, $c_k$ and a partially observable non-cost component. Executing an action \(a_k \in A\) results in a transition to a new state \(s_{k+1} \in S\) at the next time step with probability \(T(s_k, a_k, s_{k+1})\). In this, the transition for the cost component is deterministic with $c_{k+1}=c_k+ C(a_k)$. 

The agent receives an observation \(o_k \in \Omega\) about the environment state with probability \(O(s_{k+1}, a_k, o_k)\), dependent on \(s_{k+1}\) and \(a_k\). The agent, not having direct access to its true non-cost component, updates its belief using the observation \cite{kaelbling1998planning}.
A reward \(r_k = R(s, a)\) is received for the action taken.

The objective in optimizing a policy for a finite-horizon b-POMDP is to identify a sequence of actions that maximizes the total expected reward over the planning horizon while adhering to the total budget $B$. 

The budget constraint can be formulated as
 \begin{equation}
 \label{eqn2}
 0 \leq c_k \leq B \text{ } \forall k \in \{0, 1, \cdots, H\} . 
 \end{equation}

\section{Problem Formulation}
\label{probf}
This work addresses the problem of optimal trajectory tracking for an agent acting within the structure of a b-POMDP with an active localization budget, which represents the maximum allowable number of active localization updates. The agent's state, denoted as \(s\), is composed of two key elements:
\begin{itemize}
    \item \(s^p\): A partially observable component representing physical attributes of the agent.
    \item \(s^l\): A fully observable component indicating the remaining budget.
\end{itemize}
The transition function governing is given by:
\begin{equation*}
    s_{k+1} = \begin{bmatrix} s_{k+1}^{p} \\ s_{k+1}^l \end{bmatrix} = \begin{bmatrix} f^{p}(s_k^{p}, u_k^p, w_k) \\ s_k^l - u_k^l \end{bmatrix},
\end{equation*}
where \(s_k^p\) and \(s_k^l\) identify the state components at time step \(k\). The control input at each time step, \(u_k\), is divided into \(u_k^p\) for actions that affect the physical state $s_k^p$ and \(u_k^l\) for active update decisions. Here, \(u_k^l\) is a binary variable, with 0 denoting a passive localization update and 1 signifying an active update. Also, the physical action component $u^p$ is bounded, $u^p \in [u^{p,\min}, u^{p,\max}]$.
The term \(w_k\) denotes the zero-mean stochastic disturbance at time step \(k\) affecting \(s_{k+1}^p\). The cost function for our problem is defined as $C(u_k)=u_k^l$

Observations regarding $s_k^p$ are determined by the observation function:
\begin{equation}
    y_k = u_k^l \cdot s_k^p. \label{obs}
\end{equation}
Model (\ref{obs}) implies that the agent acquires its true state if it opts for an active localization update (\(u_k^l=1\)); otherwise, it receives no observation (\(y=0\)). Consequently, while \(s_k^l\) is always fully observable, the agent must rely on its belief about \(s_k^p\) to navigate and make informed decisions in scenarios devoid of observations.

\subsection{Problem Statement}
The core objective of this work is to develop an optimal policy for this b-POMDP with horizon $H$ and total active localization budget $B_l$, aiming to fulfill two requirements: 
\begin{enumerate}
    \item ensuring that the agent tracks a given reference trajectory with high precision, and 
    \item it does so while adhering to the budget.
\end{enumerate}

Requirement (\ref{eqn2}) challenges the agent to optimize its trajectory following capabilities under partial observability while strategically leveraging limited active localization updates to access its true state, thus striking a balance between trajectory tracking efficacy and optimal resource management.

\section{Solution Approach}
In this section, we elaborate on our approach for deriving the optimal policy for the problem outlined in Section \ref{probf}. We present ComTraQ-MPC, a novel hybrid framework that integrates meta-trained Deep Q-Networks (DQN) with Model Predictive Control (MPC) to address the challenges inherent in solving the b-POMDP. 

\subsection{ComTraQ-MPC}
ComTraQ-MPC provides a framework to compute optimal policies for the b-POMDP described earlier. Fig. \ref{fig:arch} illustrates the architecture of this framework. 
Distinctively, owing to the large action space, ComTraQ-MPC deconstructs the decision-making problem into two specialized sub-problems: trajectory tracking and active localization.
Doing so allows the method to leverage the proven strengths of MPC for trajectory tracking and the adaptive decision-making capabilities of DQN for determining optimal active update policies. 
Through ComTraQ-MPC, we propose an integration of MPC and DQN to address the dual challenges of precise trajectory tracking and resource management. 

\subsubsection{MPC for Trajectory Tracking}

MPC optimizes future control actions over a finite horizon to minimize a predefined cost function \cite{garcia1989model}. In our framework, MPC is employed to control the \(s^p\) component of the state. 
Due to the partial observability of \(s^p\), the MPC utilizes the mean belief of this component, \(\bar{s}^p\), to guide the trajectory tracking process. This average belief is estimated using a particle filter approach (passive localization) \cite{coquelin2008particle}.

We formulate the MPC optimization problem as:
\begin{equation}
\label{opt}
\begin{aligned}
\min_{u_{k:k+H_{MPC}}} & \sum_{i=k}^{k+H_{MPC}} \| \bar{s}^p_i - s^{\text{ref}}_i \|^2, \\
\text{subject to} & \quad \bar{s}_{k+1}^p = f^p(\bar{s}_k^p, u_{k}^p), \text{ } \forall k,
\end{aligned}
\end{equation}

where \(s^{ref}_{i}\) is the state of the reference trajectory at the same step and $H_{MPC}$ is the prediction horizon. This cost function aims to minimize the distance between the agent's estimated state and the reference trajectory, thus ensuring precise trajectory tracking.
The control action \(u_k^p\) is determined as part of the MPC optimization process.


\subsubsection{DQN for Localization Decisions with Meta-Training}

Within the ComTraQ-MPC framework, DQN is employed to optimize localization decisions under the constraints of limited active localization updates. The introduction of meta-training extends the capability of DQN to adapt across a diverse set of trajectories, enhancing its generalizability.

The state for the DQN, \(\bar{s}_k\), includes the agent's average belief about its physical state, \(\bar{s}_k^p\), the variance of this belief, $\sigma_{s_k^p}$, and the remaining active localization budget, \(s_k^l\). The goal is to judiciously use active localization updates, \(u_k^l \in \{0, 1\}\), that balance the necessity of accessing the agent's true state against the cost of actively localizing.

Meta-training modifies the traditional DQN training process by introducing a loss function that accounts for the performance across multiple trajectories and budgets, aiming to minimize the expected loss over a distribution of tasks:
\begin{multline*}
    L(\theta) = \mathbb{E}_{\substack{T^{ref} \sim \mathcal{T} \\ B_c \sim \mathcal{B}}}\left[ \mathbb{E}_{\bar{s}_k, u_k^l} \left[ \left( Q_{T^{ref}}(\bar{s}_k, u_k^l | \theta) - y_{T^{ref}} \right)^2 \right] \right].
\end{multline*}
Here \(\mathcal{T}\) denotes the distribution of trajectories, \(T^{ref}\) a specific trajectory from this distribution. Similarly, $\mathcal{B}$ and $B_c$ represent the distribution of active localization budgets and specific budget from this distribution. The target \(y_{T^{ref}}\) is defined as \(y_{T^{ref}} = r_{T^{ref}} + \gamma \max_{u_{k+1}^l} Q_{T^{ref}}(\bar{s}_{k+1}, u_{k+1}^l | \theta^-)\), with \(r_{T^{ref}}\) being the immediate reward under trajectory \(T^{ref}\), budget $B_l$, and \(\theta^-\) representing the parameters of a target network for stability.

This loss function ensures that the DQN learns a policy not just for a single trajectory and active localization budget but is robust across the variations encountered in different scenarios. 

\subsubsection{Interplay between DQN and MPC for a given reference trajectory and active localization budget}
The key component of the ComTraQ-MPC framework is the interaction between the two modules (DQN and MPC).

\begin{figure*}[th]
    \centering
    \begin{subfigure}{0.49\textwidth}
        \centering
        \includegraphics[width=0.48\linewidth, clip, trim=2cm 1.5cm 12cm 1.5cm]{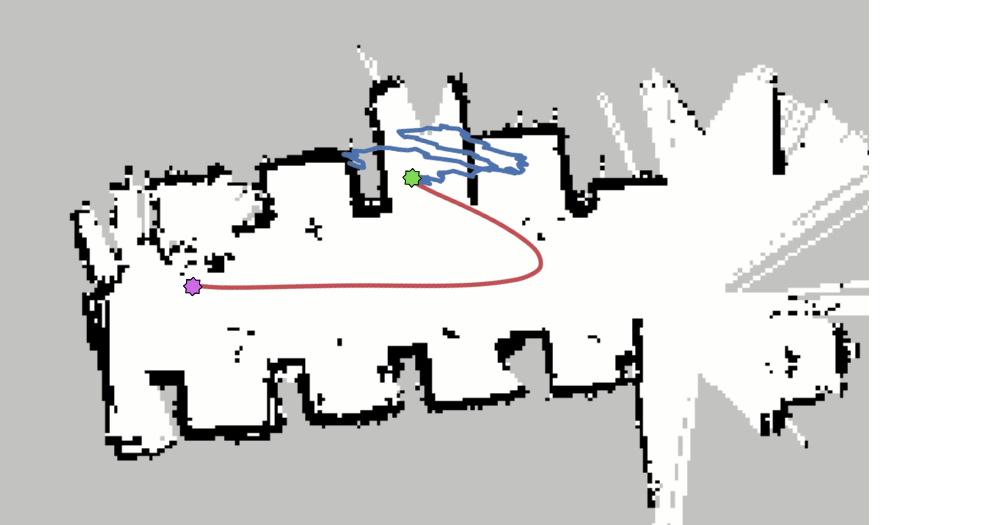}
        \hfill
        \includegraphics[width=0.48\linewidth, clip, trim=0 0 12cm 0]{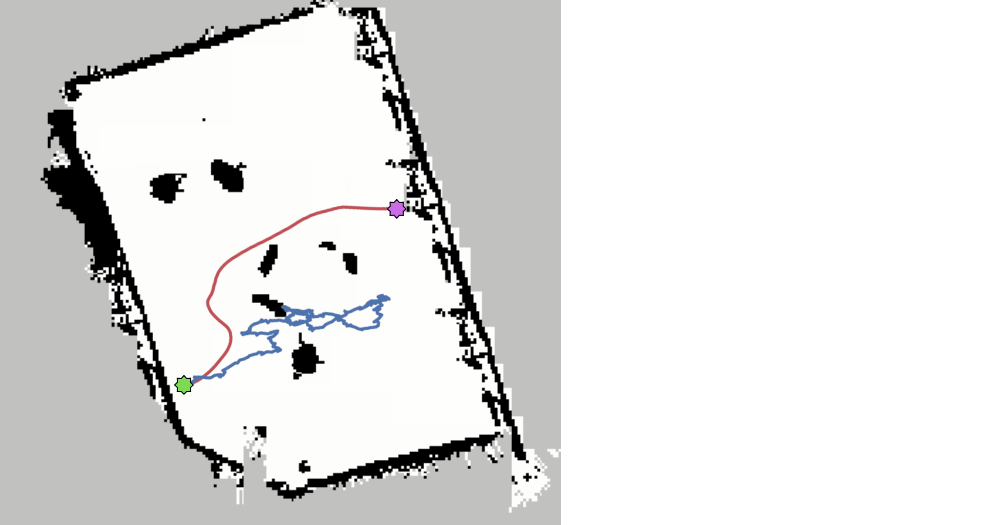}
        \caption{Trajectory tracked by MPC with passive localization}
        \label{fig:pair1}
    \end{subfigure}%
    \vspace{0.1cm}
    \hfill
    \begin{subfigure}{0.49\textwidth}
        \centering
        \includegraphics[width=0.48\linewidth, clip, trim=2cm 1.5cm 12cm 1.5cm]{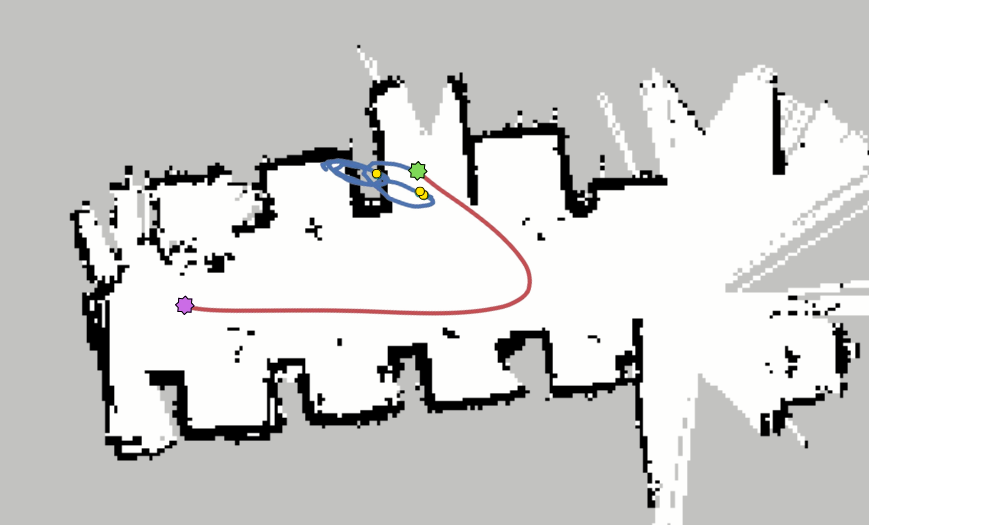}
        \hfill
        \includegraphics[width=0.48\linewidth, clip, trim=0 0 12cm 0]{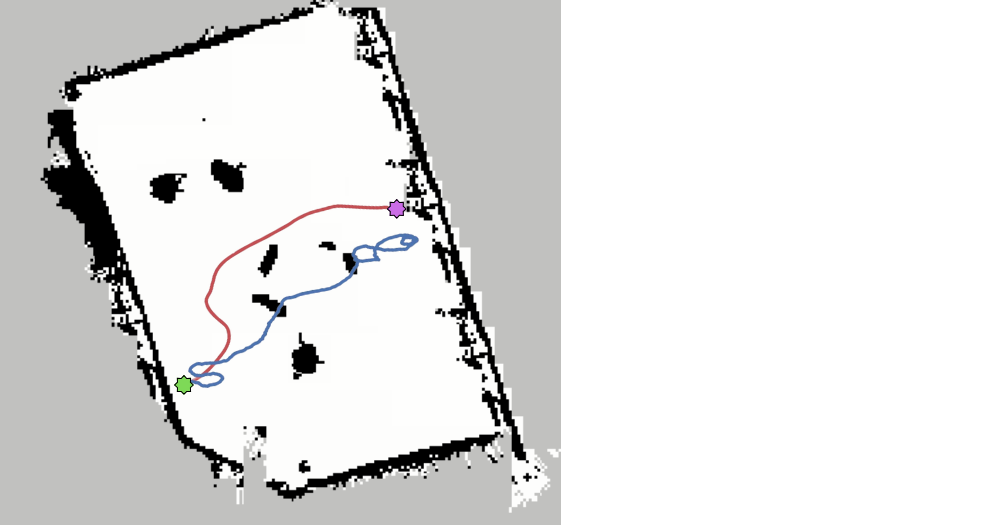}
        \caption{Trajectory tracked by DQN}
        \label{fig:pair2}
    \end{subfigure}
    \vspace{0.1cm}
    
    \begin{subfigure}{0.49\textwidth}
        \centering
        \includegraphics[width=0.48\linewidth, clip, trim=2cm 1.5cm 12cm 1.5cm]{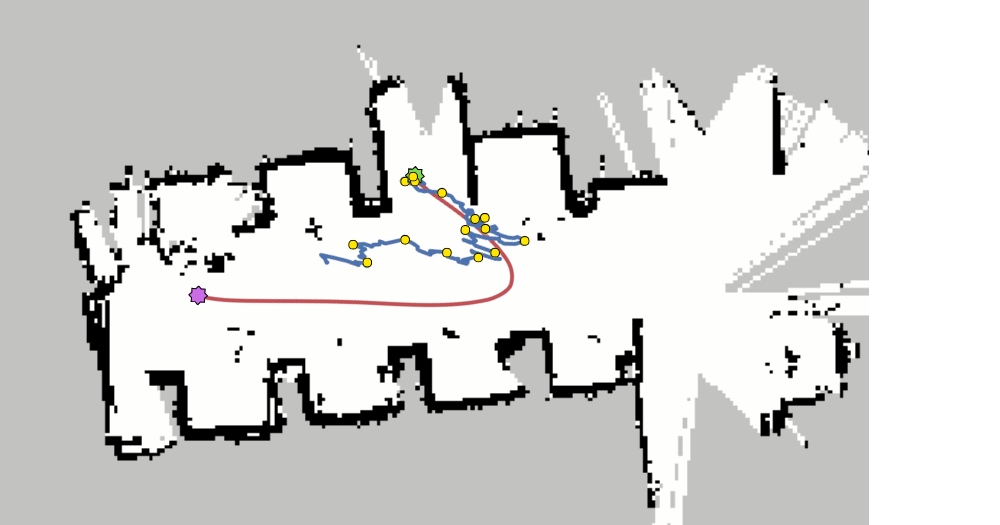}
        \hfill
        \includegraphics[width=0.48\linewidth, clip, trim=0 0 12cm 0]{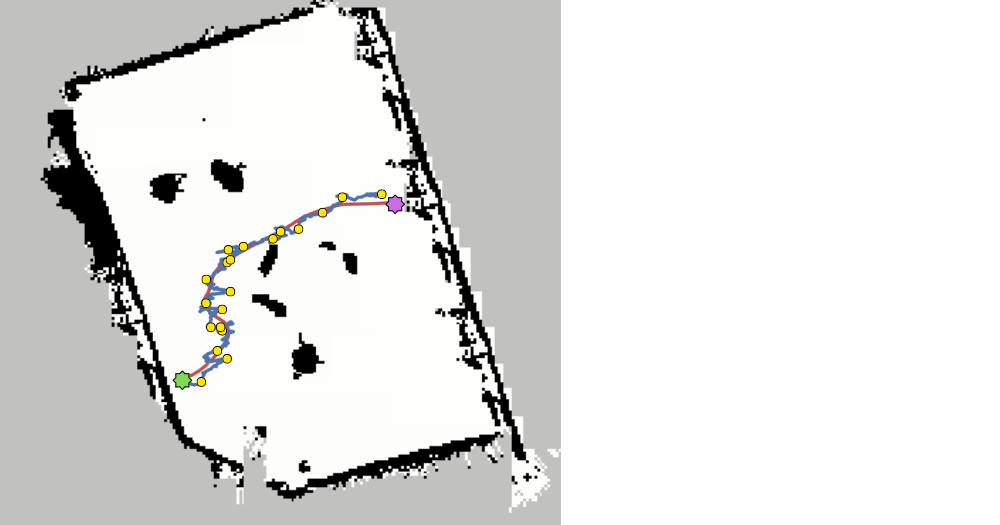}
        \caption{Trajectory tracked by MPC with naive localization policy}
        \label{fig:pair3}
    \end{subfigure}%
    \hfill
    \begin{subfigure}{0.49\textwidth}
        \centering
        \includegraphics[width=0.48\linewidth, clip, trim=2cm 1.5cm 12cm 1.5cm]{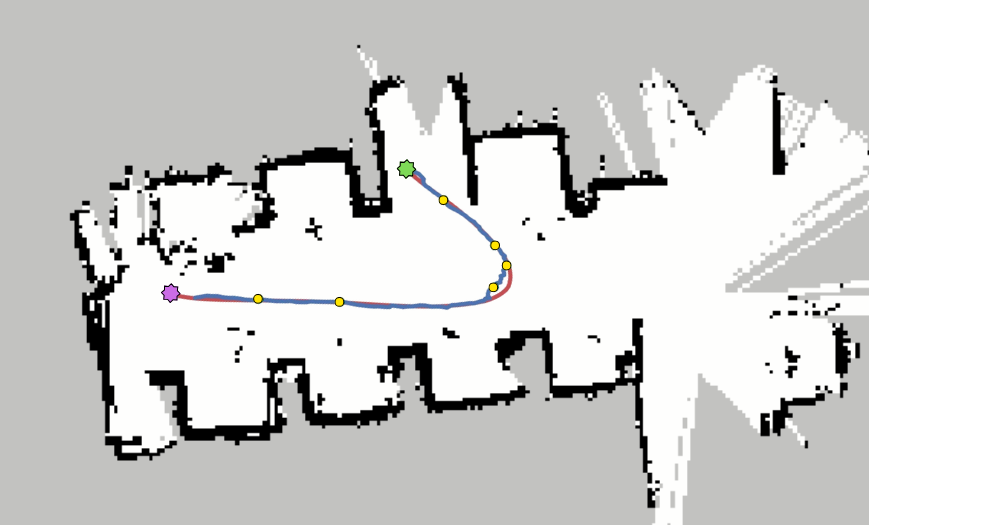}
        \hfill
        \includegraphics[width=0.48\linewidth, clip, trim=0 0 12cm 0]{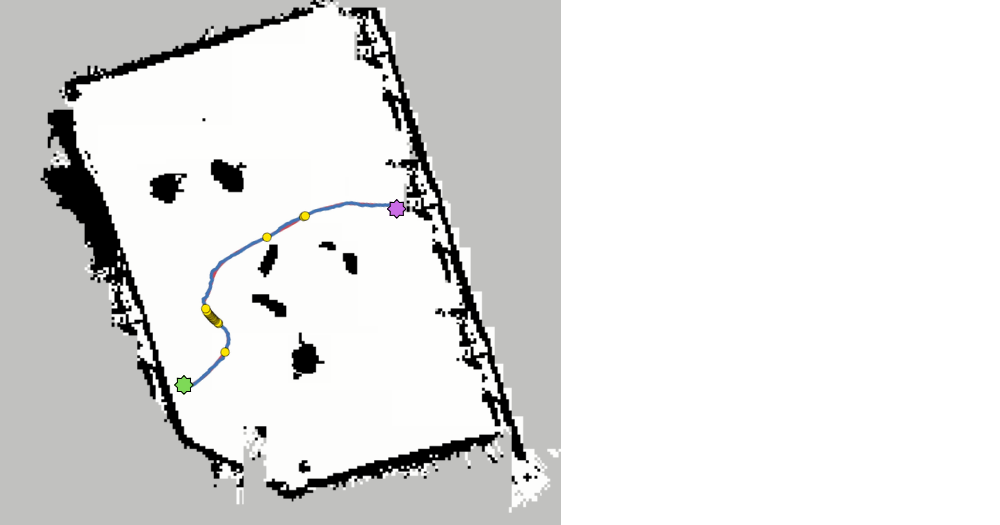}
        \caption{Trajectory tracked by  ComTraQ-MPC}
        \label{fig:pair4}
    \end{subfigure}
    
    \caption{Comparison of trajectory tracking across scenarios. In each subfigure, the left image illustrates the trajectory in Scenario~1, while the right image depicts Scenario~2. Key points are color-coded: \textcolor{green}{green} for the \textcolor{green}{start}, \textcolor{DarkOrchid}{purple} for the \textcolor{DarkOrchid}{goal}, \textcolor{yellow}{yellow} for \textcolor{yellow}{active localization updates}. The \textcolor{red}{reference trajectory} is shown in \textcolor{red}{red}, and the trajectory produced by the \textcolor{blue}{evaluated approach} is in \textcolor{blue}{blue}. }
    \label{fig:paired_images}
\end{figure*}

\paragraph{Influence of DQN on MPC}
DQN determines the optimal localization action $u_k^l$ based on its state \(\bar{s}_k\), which influences the MPC's belief state estimate and subsequent control actions. The localization decision updates the state estimate \(\bar{s}_{k+1}^p\) for MPC, as:
\begin{equation}
    \bar{s}_{k+1}^p = \begin{cases} 
    s_{k+1} & \text{if } u_k^l = 1, \\
    f^p(\bar{s}_k^p, u_{k}^p, w_k) & \text{otherwise}.
    \end{cases}
\end{equation}
This decision directly impacts the MPC's trajectory optimization by modifying the certainty level of state information -- leading to adjustments in the control strategy to solve the optimization problem given in (\ref{opt}).

\paragraph{Feedback from MPC to DQN}
The deviation from the desired trajectory, as determined by MPC, plays a crucial role in refining the decision-making process of DQN. This relationship is formalized through a feedback mechanism where the performance of MPC provides a learning signal \(r_{\text{MPC}}\) to DQN. This signal quantifies the control efficacy of MPC given the prevailing localization strategy:
\begin{equation}
    r_{\text{MPC}} = -\alpha \sum_{i=k}^{k+H} \| \bar{s}_i^p - s^{ref}_i \|^2,
\end{equation}
where \(\alpha\) serves as a weighting parameter, modulating the influence of trajectory adherence on the learning process. This feedback incentivizes DQN to prioritize localization decisions that contribute to minimizing trajectory deviations.

The dynamic exchange between DQN's policy for localization and the control outputs of MPC establishes a feedback loop, crucial for the continual enhancement of the ComTraQ-MPC framework's adaptability and efficacy. The augmented reward function for DQN is  represented as:
\begin{equation}
    r_{DQN}=\textstyle \begin{cases}
     r_{\text{MPC}} - (1-\alpha) \| s_k^p - \bar{s}_k^p \|, & \text{if } 0 \leq s_k^l \leq B_l \\
    R_{\text{min}}, & \text{otherwise} 
\end{cases} 
\label{rewards}
\end{equation}

where \(r_{\text{MPC}}\) is the aggregated control performance and \(r_{deviation}=(1-\alpha)\|s_k^p - \bar{s}_k^p\|\) accounts for the deviation between the actual state \(s_k^p\) and the estimated state \(\bar{s}_k^p\). 
We only need access to true state data while training the DQN and not during actual mission performance.
$R_{\text{min}}$ is a large negative reward for exceeding the active localization budget. 

\section{Experimental Evaluation}

This section evaluates the efficacy of the ComTraQ-MPC framework through a comprehensive evaluation in the context of autonomous ground vehicle trajectory tracking, where a trajectory is defined as an ordered set of closely spaced waypoints. First, we describe the b-POMDP framework that models the system's physical dynamics as given in \cite{polack2017kinematic} and the localization update dynamics as described in Section \ref{probf}. Subsequently, we compare ComTraQ-MPC's performance against established baselines. This comparison spans both simulated environments and real-world hardware implementations.

Hardware experimentation is carried out utilizing a TurtleBot Burger, designed to undertake the task of tracking a pre-defined reference trajectory obtained using the advanced navigation package within ROS. The TurtleBot is fitted with an single-board computer and utilizes a LIDAR sensor for SLAM \cite{1638022}. 
We also leverage ROS to interface the sensor with the onboard computer.

\textbf{State:}
We model the agent as a b-POMDP with the state at time $k$ given by: 
\[
s_{k} = \begin{bmatrix} s_{k}^{p} \\ s_{k}^l \end{bmatrix} = \begin{tikzpicture}[baseline=(current bounding box.center)]
\matrix (m) [matrix of math nodes,left delimiter=[,right delimiter={]}] {
    x_k \\
    y_k \\
    v_k \\
    \psi_k \\
    b_k \\
};
\draw[decorate,decoration={brace,amplitude=5pt,raise=1pt},thick] ($(m-1-1.north east)+(0.5,0)$) -- ($(m-4-1.south east)+(0.5,0)$) node[midway,xshift=0.95cm] {$s_k^p \in \mathbb{R}^4$};
\draw[decorate,decoration={brace,amplitude=5pt,raise=1pt},thick] ($(m-5-1.north east)+(0.5,-0.05)$) -- ($(m-5-1.south east)+(0.5,-0.05)$) node[midway,xshift=0.95cm] {$s_k^l \in \mathbb{N}_0$};
\end{tikzpicture}.
\]
Here, $s_k^p$ describes the position ($[x_k, y_k]$), velocity ($v_k$) and orientation ($\psi_k$) of the agent and $s_k^l$ describes the remaining budget for active localization ($b_k$).


\begin{figure*}[th]
    \centering
    \includegraphics[width=0.9\textwidth]{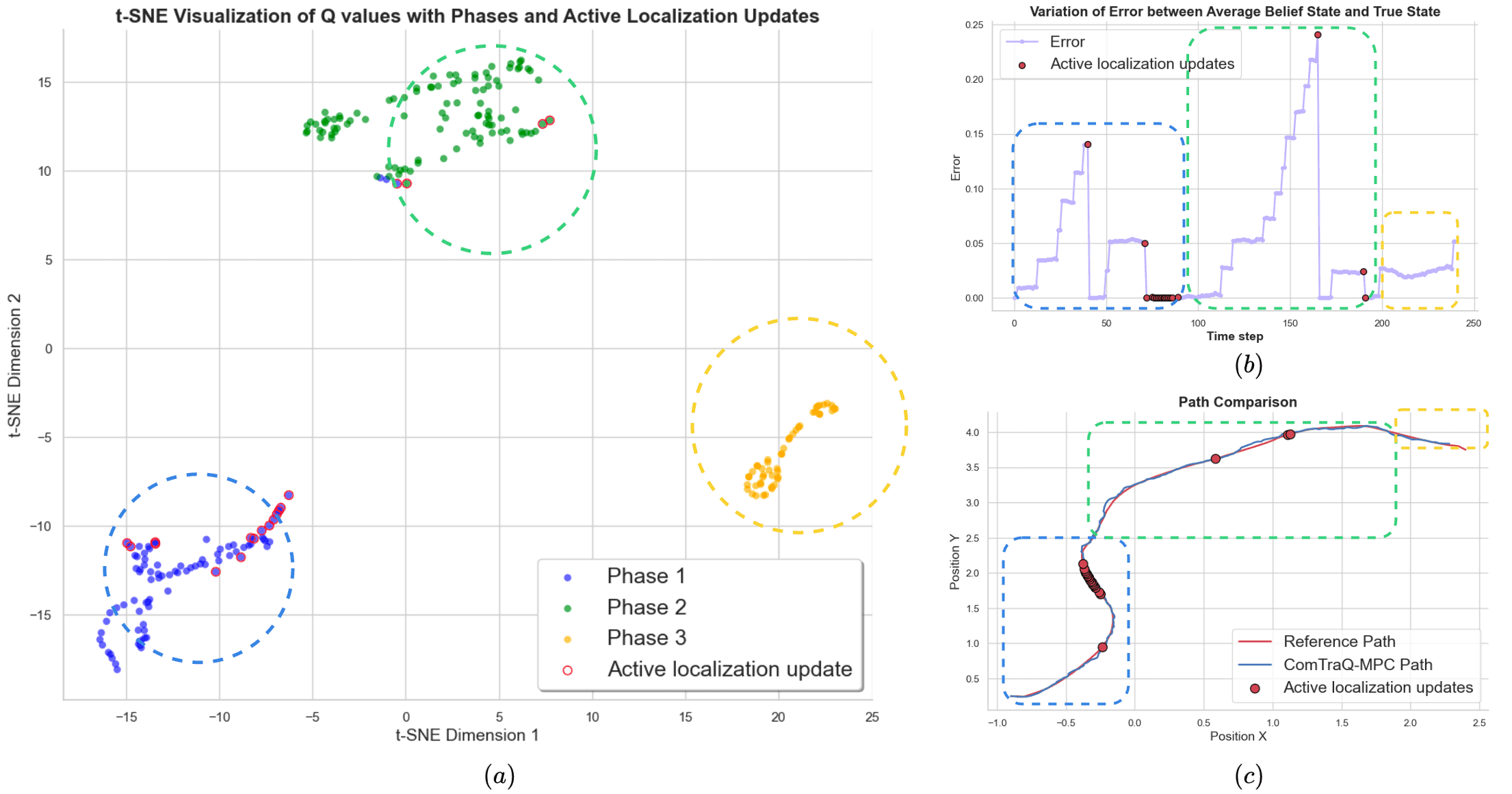}
    \caption{Integrated Analysis of ComTraQ-MPC in Scenario~2. (a) t-SNE visualization of Q-values, highlighting the distinct clusters corresponding to \textcolor{blue}{Phase~1 (blue)}, \textcolor{ForestGreen}{Phase~2 (green)}, and \textcolor{yellow}{Phase~3 (yellow)} of the mission, with active localization updates denoted by \textcolor{red}{red circles}. (b) Variation of error between the average belief state and the true state, with active localization updates superimposed, corresponding to the color-coded mission phases in (a). (c) Path comparison illustrating the reference path and the ComTraQ-MPC path with active localization updates marked, mirroring the sequence of the mission phases as color-coded in (a) and (b).}
    \label{fig:tsne-final}
\end{figure*}
\textbf{Actions:}
The dynamics of the agent are governed through two distinct types of actions: physical action, \(u_k^p\), and  localization action, \(u_k^l\). Specifically, \(u_k^p\) encompasses acceleration (\(a\)) and steering angle (\(\delta\)), with the acceleration bounded within \([-0.2, 0.2]\) m/s\(^2\) and the steering angle constrained to \([-60^\circ, 60^\circ]\). 

\textbf{Partial Observability of $\mathbf{s_k^p}$:}
The agent operates under conditions of partial observability, receiving observations of its physical state when it actively localizes. 
These observations are derived from the agent's interaction with a SLAM-generated map, adhering to the observation function outlined in (\ref{obs}).

\textbf{Model Uncertainty:}
To emulate slip in wheels, we implement an adversarial controller that modulates the wheel torque \cite{puthumanaillam2023weathering}. Owing to this slip, the orientation of the agent has an additional deviation (in degrees) $w \sim \mathcal{N}(0,\sigma^2)$ where $\sigma = 15^\circ$. The effect of this slip combined with the partial observability can be seen in Fig. \ref{fig:pair1} where we use an MPC approach which assumes noiseless dynamics and relies only on passive localization to track a trajectory.

\textbf{Meta-Training of DQN: }
The meta-training procedure is instrumental in ensuring that the DQN component acquires a versatile and robust policy capable of adapting to a wide range of reference trajectories and active localization budgets, including those not encountered during the training phase.
This meta-training was conducted across a set of 100 feasible randomly trajectory-budget pairs. The budget allocated was proportional to the trajectory length. 

\subsection{Simulation and Hardware Results}

\textbf{Baselines:}
To evaluate the efficacy of our framework comprehensively, we benchmark its performance against well-defined baselines, each representing a distinct approach to the problem of trajectory tracking and localization in stochastic, partially observable conditions. The baselines include:
\begin{table*}[!htb]
    \centering
    \renewcommand{\arraystretch}{1.2} 
    \begin{tabular}{c | c c c | c c c }
    \arrayrulecolor{black}\hlineB{2.5}
    \rowcolor{Gray}
    & \multicolumn{3}{c|}{\begin{tabular}[c]{@{}c@{}}\textbf{Scenario 1}\\ Path length: $135$, $B_l: 10$\end{tabular}} & \multicolumn{3}{c}{\begin{tabular}[c]{@{}c@{}}\textbf{Scenario 2}\\ Path length: $245$, $B_l: 20$\end{tabular}} \\
    \arrayrulecolor{black}\clineB{2-7}{2.5}
    \rowcolor{LightGray}
    & \textbf{\# waypoints followed} & \textbf{MAE}  & \textbf{Goal reached} & \textbf{\# waypoints followed} & \textbf{MAE} &  \textbf{Goal reached} \\
    \arrayrulecolor{black}\hlineB{2.5}
    MPC with passive localization & 23 & 2.748 & no & 9 & 1.485 & no \\
    \hline
    DQN & 28 & 0.990 & no & 26 & 2.341 & no \\
    \hline
    MPC with naive localization policy & 51 & 0.904 & no & 119  & 0.131 & yes  \\
    \hline
    \textit{\textbf{ComTraQ-MPC}} & \textbf{ 105 }& \textbf{0.276} & \textbf{yes} & \textbf{242} & \textbf{0.030} & \textbf{yes}  \\
    \arrayrulecolor{black}\hlineB{2.5}
    \end{tabular}
    \vspace{5mm} 
    \caption{Comparative Analysis of Navigation Performance: This table compares the number of waypoints followed (within a radius of 0.1m), mean absolute error (MAE), and whether the goal was reached or not. Successful goal attainment is defined as the TurtleBot arriving within a 0.1m radius of the target.}
    \label{tab:comparison}
\end{table*}
\begin{itemize}
    \item \textbf{MPC with passive localization:} This explores the performance of MPC in trajectory tracking without the benefit of active localization updates. It serves to highlight the impact of active localization on tracking accuracy and the inherent limitations when operating solely based on initial state information.
    \item \textbf{Learning-based strategies:} This baseline employs vanilla DQN to manage both trajectory tracking and localization decisions. It showcases the capabilities of a purely reinforcement learning-based approach in navigating the complexities of the task without the predictive advantage of MPC.  
    \item \textbf{MPC with naive localization policy:} Contrasting with the adaptive localization strategy of ComTraQ-MPC, this baseline implements MPC with a naive localization policy, where active localization occurs at predefined, regular intervals---testing the effectiveness of periodic updates versus the dynamic decision-making process of ComTraQ-MPC.
\end{itemize}

\textbf{Scenarios:}
Our empirical analysis comprises experiments conducted in two distinct trajectory-budget pairs to assess the adaptability and performance of the ComTraQ-MPC framework. The first pair is one that the agent has previously encountered during its meta-training phase (refer to Fig. \ref{fig:teaser1}), while the second represents an unseen trajectory, unexperienced by the agent prior to testing (see Fig. \ref{fig:teaser2}). 

\underline{Scenario~1 -- Previously Encountered Trajectory:} This scenario evaluates ComTraQ-MPC's performance on a trajectory that is a part of its meta-training dataset. The familiarity of the trajectory and active localization budget allows us to examine the efficiency of the learned localization and navigation strategies under conditions that the system is optimally prepared for. The trajectory is relatively a short one with 135 waypoints and the agent has an active localization budget of 10 for tracking it. 

\underline{Scenario~2 -- Unencountered Trajectory:} Contrasting with Scenario 1, this experiment presents the agent with an entirely unseen trajectory-budget pair, not included in the meta-training process. This scenario serves to test the generalizability of ComTraQ-MPC, probing its capacity to apply learned policies to new conditions using the previously meta-trained model. The unencountered trajectory is longer in length, comprising of 245 points and a higher active localization budget of 20. 

\subsection{Discussions}
Our experiments reveal the inadequacies of conventional methods in navigation tasks under partial observability and active localization limits, highlighted by the failure of the passively localized MPC (Fig. \ref{fig:pair1}). The DQN strategy, even after extensive training, often neglects the strategic value of active localization (Fig. \ref{fig:pair2}), likely due to an expansive action space that combines motion and localization actions. 

Fig. \ref{fig:pair3} depicts the failure of the naive MPC strategy, which uses uniformly spaced active localization intervals determined by dividing the trajectory length by the active localization budget. This underscores the necessity for a dynamic approach to active localization update, capable of responding to the path's evolving challenges.
 


Table \ref{tab:comparison} compares the number of waypoints followed in both the scenarios when different approaches are used. As can be observed (see Fig. \ref{fig:pair4}) ComTraQ-MPC not only follows the maximum number of waypoints but also does so with the least trajectory tracking error (MAE). The performance disparity of all the approaches (except DQN) across scenarios can be attributed to the magnified effect of slip on the shorter trajectory in Scenario~1. The anamolous performance of DQN might be due to the smaller training state space in Scenario~1 in comparison to Scenario~2. Finally, our framework's performance on Scenario~2 underscores its adaptability to unseen scenarios.

\subsubsection{Analysis of ComTraQ-MPC on Previously Unseen Trajectory-Budget Pairs}
In Scenario~2, which presents a trajectory-budget pair not encountered during the meta-training phase, ComTraQ-MPC demonstrates its robustness and adaptability by achieving optimal trajectory tracking, significantly surpassing baseline methods as illustrated in Scenario~2 of Fig. \ref{fig:pair4}. 

Fig. \ref{fig:tsne-final} provides an analysis into the decision-making process of ComTraQ-MPC in this scenario. We split the trajectory into 3 major phases, Phase~1 (from start point to the second major turn of the trajectory), Phase~2 (from second major turn to final turn) and Phase~3 (from final turn to goal). It reveals that ComTraQ-MPC adopts an adaptive localization strategy that intelligently modulates based on the mission's phase. Specifically, in Phase~1 of Scenario~2, characterized by sharper turns, ComTraQ-MPC strategically increases active localization updates between the first and second major turns of the trajectory. 

Conversely, during Phase~2, where the trajectory is less complex as compared to Phase~1, the number of active localization updates significantly reduces. 
This reduction is predicated on the agent's assessment that fewer active updates are necessary for successful navigation as it approaches the final phase.
Finally in Phase~3, where the goal proximity becomes more apparent, ComTraQ-MPC chooses not to actively localize. 

This nuanced approach to localization can also be visualized in the t-SNE \cite{van2008visualizing} analysis of Q-values in Fig. \ref{fig:tsne-final}a. The clustering of Q-values indicates distinct decision-making patterns: states in Phase~1 are grouped together (bottom left, marked by a blue circle), suggesting a higher propensity for active localization; Phase~2 exhibits a different pattern (top, marked by a green circle), and Phase~3 is distinctly clustered (right, marked by a yellow circle), reflecting the strategic reduction in active localization updates.


\section{Conclusions}

This paper introduced ComTraQ-MPC, a novel framework that integrates Deep Q-Networks (DQN) with Model Predictive Control (MPC) to address the intricate challenge of trajectory tracking in partially observable, stochastic environments constrained by limited active localization updates. Our approach uniquely combines the adaptive decision-making prowess of DQN, enhanced through meta-training across diverse trajectory-budget pairs, with the precision and foresight of MPC for effective trajectory planning and execution.
The empirical evaluations conducted in both previously seen and unseen scenarios demonstrate the superior performance of ComTraQ-MPC over traditional methods. By optimizing the use of the limited active localization updates and efficiently tracking reference trajectories, our framework not only achieves significant improvements in operational efficiency and accuracy but also exhibits remarkable adaptability to a wide range of scenarios. 
Future work will focus on extending the framework's capabilities to more complex multi-agent scenarios to deal with active localization along with communication amongst the agents.

\section*{Acknowledgement}
We would like to thank Leonardo Bobadilla, Paulo Padrão, and José Fuentes from Florida International University for their insights and discussions leading to this paper.

\bibliographystyle{IEEEtran}
\bibliography{references}

\end{document}